%% file: warpdiffusion_arxiv.tex
\documentclass[10pt,twocolumn,letterpaper]{article}

\usepackage[pagenumbers]{cvpr} 

\input{preamble}

\definecolor{cvprblue}{rgb}{0.21,0.49,0.74}
\definecolor{red}{rgb}{.9,0.1,0.1}
\usepackage[pagebackref,breaklinks,colorlinks,citecolor=cvprblue]{hyperref}

\newif\ifcomments
\commentstrue 

\def\paperID{4254} 
\def\confName{CVPR}
\def\confYear{2024}

\title{WarpDiffusion: Efficient Diffusion Model for High-Fidelity Virtual Try-on}

\author{
Xujie Zhang{$^{1}$},
~ Xiu Li{$^{2}$},
~ Michael Kampffmeyer{$^{3}$},
~ Xin Dong{$^{2}$}\\ \vspace{-16pt}\\ 
Zhenyu Xie{$^{1}$},
~ Feida Zhu{$^{2}$},
~ Haoye Dong{$^{4}$},
~ Xiaodan Liang{$^{*1}$}\\\vspace{-10pt}\\
{$^{1}$}Shenzhen Campus of Sun Yat-Sen University, {$^{2}$}ByteDance \\ \vspace{-16pt}\\
{$^{3}$}UiT The Arctic University of Norway,
{$^{4}$}Carnegie Mellon University\\
\small{\tt{\{zhangxj59,xiezhy6\}@mail2.sysu.edu.cn,lixiulive@hotmail.com,michael.c.kampffmeyer@uit.no}}\\ 
\small{\tt{dongxin.1016@bytedance.com,zhufeida@connect.hku.hk,donghaoye@cmu.edu},xdliang328@gmail.com}
\vspace{-2mm}
}

\begin{document}
\twocolumn[{%
\renewcommand\twocolumn[1][]{#1}%
\maketitle
\includegraphics[width=1\hsize]{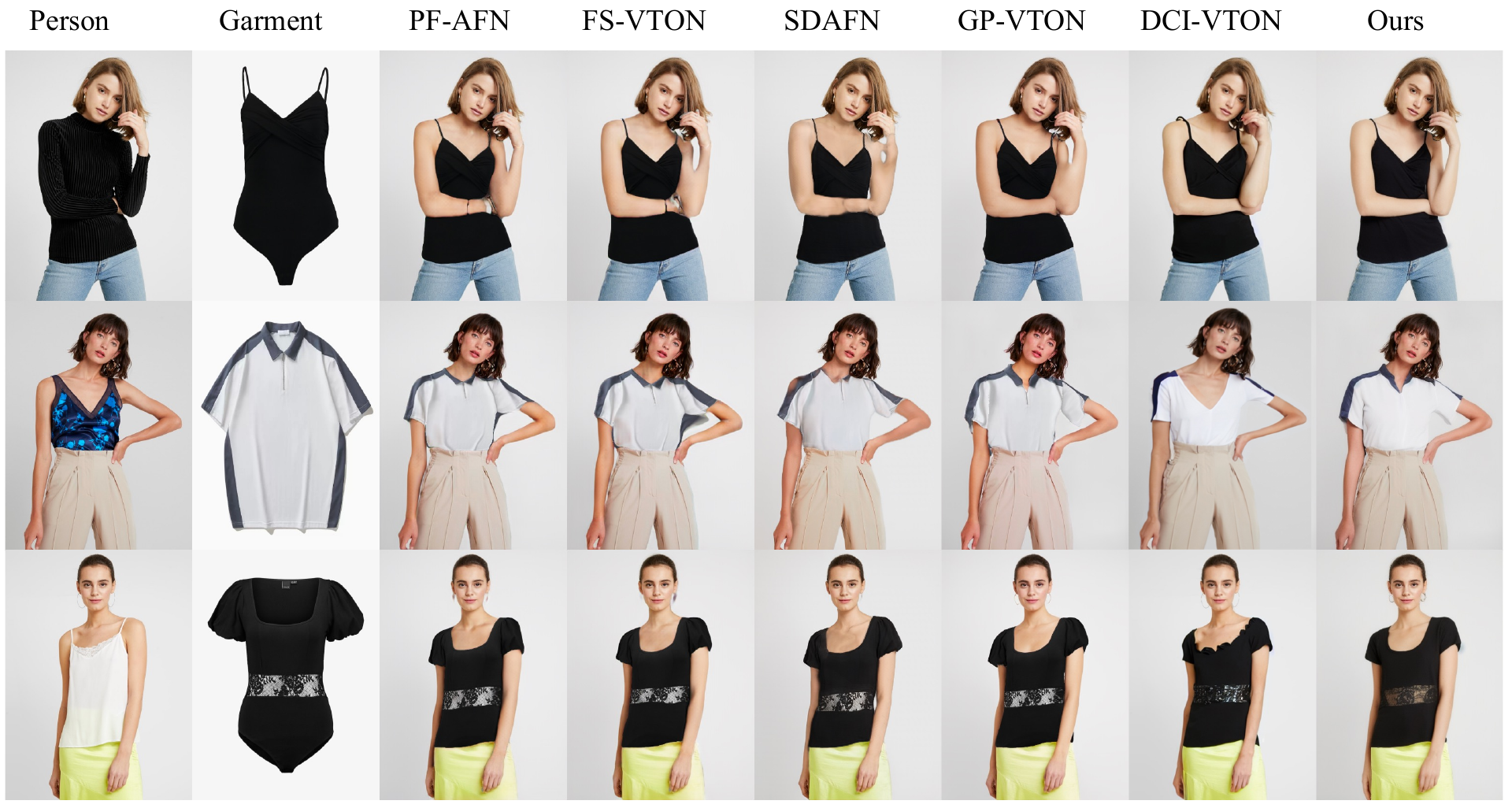}
    \captionof{figure}{WarpDiffusion surpasses current state-of-the-art methods (PF-AFN~\cite{ge2021parser}, FS-VTON~\cite{he2022style}, SDAFN~\cite{bai2022single}, GP-VTON~\cite{xie2023gp}, DCI-VTON~\cite{gou2023taming}), producing high-fidelity results with photo-realistic try-on effects, including wrinkles, shadows and skin in see-through areas.} 
    \label{fig:teaser}
    \vspace{3mm}
}]
\begin{abstract}
Image-based Virtual Try-On (VITON) aims to transfer an in-shop garment image onto a target person. While existing methods focus on warping the garment to fit the body pose, they often overlook the synthesis quality around the garment-skin boundary and realistic effects like wrinkles and shadows on the warped garments. These limitations greatly reduce the realism of the generated results and hinder the practical application of VITON techniques. Leveraging the notable success of diffusion-based 
models in cross-modal image synthesis, some recent diffusion-based methods have ventured to tackle this issue. However, they tend to either consume a significant amount of training resources or struggle to achieve realistic try-on effects and retain garment details. For efficient and high-fidelity VITON, we propose WarpDiffusion, which bridges the warping-based and diffusion-based paradigms via a novel informative and local garment feature attention mechanism. Specifically, WarpDiffusion incorporates local texture attention to reduce resource consumption and uses a novel auto-mask module that effectively retains only the critical areas of the warped garment while disregarding unrealistic or erroneous portions. Notably, WarpDiffusion can be integrated as a plug-and-play component into existing VITON methodologies, elevating their synthesis quality. Extensive experiments on high-resolution VITON benchmarks and an in-the-wild test set demonstrate the superiority of WarpDiffusion, surpassing state-of-the-art methods both qualitatively and quantitatively. \footnote{Xiaodan Liang is the corresponding author. This work is done during Xujie Zhang's internship in ByteDance.}
\end{abstract}

\section{Introduction}
Virtual Try-On (VITON) is gaining significant traction in the e-commerce industry due to its potential to revolutionize online shopping with virtual changing room experiences.
VITON achieves this by generating highly realistic images of a person wearing a specific garment from an in-shop garment image and a person image~\cite{han2018viton}.
However, most existing approaches fail to meet the following synthesis quality requirements of today’s e-commerce scenarios:
1) \textbf{Accurate Garment Deformation}: The provided garment should be precisely deformed to fit the body pose of the target person, while preserving garment characteristics such as shape and texture. 
2) \textbf{Realistic Try-On Effect}: The synthesized results should not only preserve the inherent characteristics of the garment but also incorporate realistic try-on effect, including wrinkles and shadows on the garment region;
3) \textbf{Consistent Visual Quality}: The visual quality of the results should be consistently high throughout the entire image, without any noticeable artifacts or blurriness in local regions such as the skin-garment boundary or neck region. 

Most existing methods~\cite{wang2018toward,yang2020towards,ge2021parser,lee2022high,xie2023gp} employ a two-stage framework for try-on synthesis. In the first stage, a neural network is used to model the explicit deformation field, via a Thin Plate Spline (TPS) transformation~\cite{bookstein1989principal} or appearance flow~\cite{zhou2016view}, for garment warping. In the second stage, Generative Adversarial Networks (GANs)~\cite{goodfellow2014generative} are used for try-on synthesis.
However, these two-stage methods heavily rely on the warping module and often fail to meet the above-mentioned synthesis quality requirements. Specifically, the explicit deformation field used in the first stage often struggles to handle complex non-rigid deformations, which can negatively impact the subsequent try-on synthesis. For instance, in the first row of Fig~\ref{fig:teaser}, when the designated person has a challenging body pose, such as crossed arms, baselines~\cite{he2022style,xie2023gp,bai2022single,ge2021parser,gou2023taming} all show excessive distortion where the arms intersect.

Besides, since the explicit deformation field directly performs warping in pixel space, the try-on appearance is solely determined by the original in-shop garment. As shown in the second row of Fig.~\ref{fig:teaser}, when the given garment is a flat image without any realistic wrinkles or shadows, these methods are unable to produce realistic results. 

Furthermore, due to the limited training data and training instability, GAN-based generators struggle to learn the specific distribution for fine details, resulting in inferior synthesis quality for certain local regions. For instance, these methods often generate blurry or jagged artifacts around the skin-garment boundary, which is evident in the third row of Fig.~\ref{fig:teaser}.  Further, none of the prior approaches are able to accurately capture the see-through nature of the garment in the belly region in this example.

Recently, in light of the significant success of diffusion models for image synthesis, some pioneering methods~\cite{huang2023composer,chen2023anydoor} have attempted to explore the application of diffusion models to VITON. However, these methods do not incorporate any deformation mechanism, which leads to spatial misalignment between the in-shop garment and body pose, ultimately resulting in a loss of garment details.
\begin{table}[t]
\def\arraystretch{1.2}
\small
\tabcolsep 6pt
\centering
\begin{tabular}{c c c c c c }
  \toprule
    \multicolumn{2}{c}{Method}                              
  & & Devices & Training time & Data \\
  \cmidrule{1-2} \cmidrule{4-6}  
  \multicolumn{2}{c}{TryonDiffusion\cite{zhu2023tryondiffusion}} 
  & & 32 TPU-v4 & 9 Days & 4M   \\
  \multicolumn{2}{c}{WarpDiffusion} 
  & & 8 A100 & 1 Day & 30K  \\ 
  \bottomrule
\end{tabular}
\caption{WarpDiffusion requires significantly less training resources than TryonDiffusion~\cite{zhu2023tryondiffusion}. }
\vspace{-6mm}
\label{tab:training}
\end{table}
To resolve this issue, TryOnDiffusion~\cite{zhu2023tryondiffusion} introduces an implicit deformation field via cross-attention layers~\cite{vaswani2017attention} to warp the input garment. However, while this approach leads to a scalable model for high-fidelity try-on synthesis, its training requires a prohibitive amount of resources, including 4 million paired data samples and thousands of GPU hours.

In order to leverage the generative capabilities of diffusion models for the VITON task in the face of limited resources, we introduce a novel effective and efficient model called WarpDiffusion in this work. WarpDiffusion combines the efficiency of explicit warping modules 
with the generative capability of pre-trained text-to-image diffusion models (i.e., StableDiffusion~\cite{rombach2022high}) to achieve high-fidelity try-on synthesis. It is worth noting that this bridging is not trivial and previous attempts fail to preserve garment details~\cite{gou2023taming,morelli2023ladi}. 
We solved this by introducing a novel informative and local garment feature attention mechanism that ensures accurate garment deformation and synthesis of realistic effects, while addressing the imperfections of the explicit warping results. Meanwhile, this mechanism introduces almost no overhead when compared to the original StableDiffusion model. As shown in Table~\ref{tab:training}, the entire model can be trained on 8 NVIDIA A100 GPUs at a resolution of 1024, with a significantly lower training consumption than that of TryonDiffusion~\cite{zhu2023tryondiffusion}. Furthermore, our WarpDiffusion is not reliant on a specific warping module and can be seamlessly integrated as a plug-and-play module into previous GAN-based try-on methods~\cite{ge2021parser,he2022style,bai2022single,xie2023gp}.

\begin{figure*}[t]
  \centering
  \includegraphics[ width=1.0\textwidth]{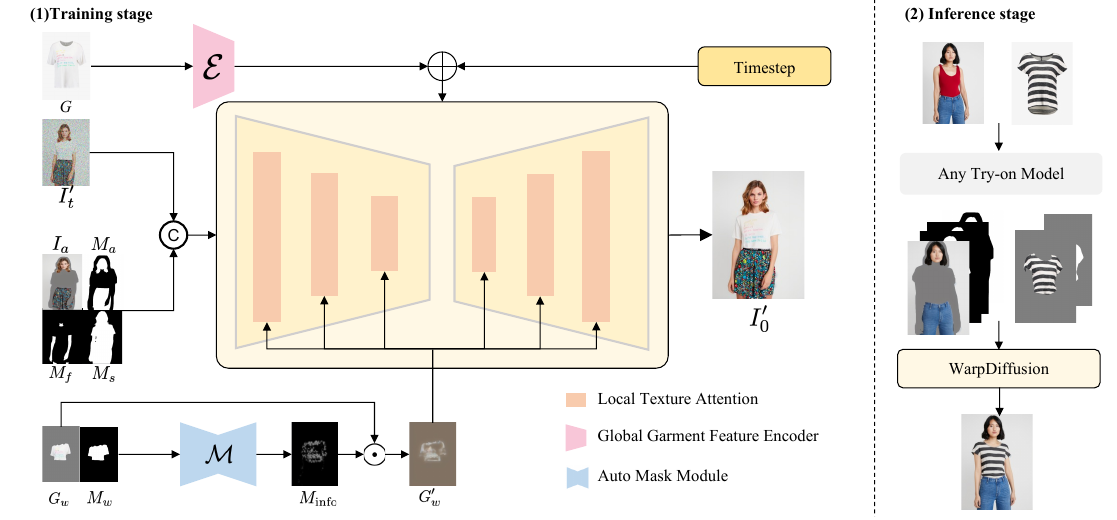}
  \caption{The framework of our WarpDiffusion. During the training stage, the garment-agnostic image $I_a$, garment-agnostic mask $M_a$, skin mask $M_s$, and foreground mask $M_f$ are concatenated with the input noise $I'_t$ and injected into the diffusion UNet. The in-shop garment $G$ is extracted via the Global Garment Feature Encoder and used in the cross-attention. Simultaneously, the warped garment $G_w$ and the warped garment mask $M_w$ are processed by the Auto Mask module to obtain an informative mask $M_\text{info}$. $M_\text{info}$ is then used to derive an improved warped garment, denoted as $G'_w$, which is subsequently utilized in the local texture attention. During the inference step, we employ the warped result from any try-on model to generate realistic try-on results, showcasing WarpDiffusion's plug-and-play capability.}
\vspace{-4mm}

  \label{fig:framework}
\end{figure*}

Overall, our contributions can be summarized as follows:
\begin{itemize}
    \item We propose WarpDiffusion, a framework integrating both explicit warping and a diffusion model, to collaboratively utilize global and local garment information for high-fidelity VITON synthesis in real-world scenarios.
    \item  WarpDiffusion leverages the pre-trained StableDiffusion and a custom-designed local texture attention to enhance the synthesis quality of the body region and significantly reduces resource consumption during training.
    \item WarpDiffusion introduces a novel Auto Mask Module for informative garment feature extraction, enabling the synthesis of realistic try-on effects such as wrinkles and shadows without loss of details.
    \item Extensive experiments conducted on public try-on benchmarks and in-the-wild test sets demonstrate that WarpDiffusion outperforms existing state-of-the-art methods and can be easily integrated as a plug-and-play module to enhance the synthesis quality of existing approaches.
\end{itemize}

\input{sec/2_related}

\input{sec/3_methods}
\input{sec/4_results}

{
    \small
    \bibliographystyle{ieeenat_fullname}
    \bibliography{main}
}


\input{sec/supp_arxiv}

\end{document}

%% file: preamble.tex
\usepackage[dvipsnames]{xcolor}


%% file: sec/2_related.tex
\section{Related work}

\textbf{GAN-Based Virtual Try-on.}
Most existing VITON methods~\cite{wang2018toward,han2019clothflow,yu2019vtnfp,dong2019towards,yang2020towards,issenhuth2020not,ge2021parser,he2022style,choi2021viton,li2021toward,chopra2021zflow,zhenyu2021wasvton,VTON_zhao2021m3d,zhenyu2021pastagan,lee2022high,bai2022single,morelli2022dress,xie2022pastagan++,dong2022wflow,zaiyu20223dgcl,xie2023gp} 
follow a conventional two-stage Generative Adversarial Network (GAN)-based pipeline for realistic try-on synthesis. In this paradigm, the first stage employs an explicit warping module to deform the in-shop garment to the target shape, while the second stage uses a GAN-based generator to fuse the deformed garment onto the reference person. The synthesis quality of these GAN-based solutions heavily relies on the deformation quality of the first stage, prompting current methods to emphasize enhancing the non-rigid deformation ability of the warping module.
However, most of the GAN-based solutions directly exploit the UNet-based~\cite{ronneberger2015u} generator for try-on synthesis and seldom explore how to improve the capability of the generator, leading to try-on results with inferior visual quality.
In this work, we mainly focus on exploring a diffusion-based generator for virtual try-on to improve the synthesis capability of existing methods in terms of the realism of the garment-skin boundary and realistic try-on effect in the garment region. 

\noindent\textbf{Diffusion-Based Virtual Try-on.}
Compared to GAN-based models, diffusion models have shown great success in high-fidelity conditional image generation~\cite{rombach2022high,saharia2022photorealistic,ramesh2022hierarchical} and editing~\cite{hertz2022prompt,mokady2023null,yang2023paint}. Image-based virtual try-on can be considered as a special case of the image editing/inpainting task conditioned on the given garment image. Therefore a straightforward way is to extend text-to-image diffusion models to accept images as the condition. 
Although methods such as~\cite{yang2023paint,huang2023composer,chen2023anydoor}
have demonstrated their ability for virtual try-on, 
these methods fail to preserve the texture detail of the try-on results because of the spatial misalignment between the in-shop garment and the reference person.

The recently proposed TryonDiffusion~\cite{zhu2023tryondiffusion} instead introduces an implicit warping mechanism for garment warping, which addresses the texture misalignment problem mentioned above and obtains promising try-on synthesis. However, the training of TryonDiffusion requires a large-scale dataset and thousands of GPU hours, which generally is prohibitive for most researchers.

Other recent approaches~\cite{morelli2023ladi,gou2023taming} aim to merge traditional GAN-based methods with diffusion models. They use the explicit warping module to create the warped garment and use the diffusion model to fuse it with the reference person image. Although these methods share a similar framework with ours, we observed that both of them fail to preserve texture details. We argue that this is due to the way they use the results of the warping module and solve this with a novel informative and local garment feature attention mechanism.

%% file: sec/3_methods.tex
\section{Methods}

\subsection{Overview}
Given an in-shop garment image $G$ and a person image $I$, image-based VITON aims to seamlessly fit $G$ onto $I$ to synthesize photo-realistic try-on result $I'$. 
To achieve this, WarpDiffusion first leverages a warping module to estimate the deformation field $F$, which is used to deform the garment image $G$ to the warped garment image $G_w$. Note that WarpDiffusion is designed to be agnostic to the choice of warping module and can thus act as a plug-and-play solution to any existing warping-based approach. Subsequently, WarpDiffusion integrates the garment image $G$, the warped garment $G_w$, and the garment-agnostic person information $I_a$\footnote{The garment-agnostic information contains the face, hair, background and all areas that are not involved in the try-on process.} to obtain the final try-on result $I'$, which not only captures texture details in the garment image, but also generates realistic try-on effects like wrinkles and shadows. 


To speed up network convergence and inherit its powerful generative ability, WarpDiffusion is built upon StableDiffusion (SD)~\cite{rombach2022high} (see Fig.~\ref{fig:framework}). More specifically, we modify the StableDiffusion inpainting model in the following ways to make it a VITON generator:
\begin{itemize}
    \item Besides the garment-agnostic image $I_a$ and the corresponding garment-agnostic mask $M_a$, we concatenate two additional features, namely the skin mask $M_s$ and the foreground mask $M_f$ 
    , with $z_T$, which is used as a hint of the try-on area ({\bf Pixel Aligned Feature Guidance})
    \item We add the global feature extracted from garment $G$ into the timestep embedding, enabling the expression of high-level garment characteristics. Simultaneously, we replace the text prompts in SD with the warped garment $G_w$. The auto mask module processes the warped garment to retain the most informative features. Subsequently, a local texture attention mechanism is employed to fuse the improved garment feature into the try-on result, enhancing texture consistency ({\bf{Garment Feature Guidance}}).
\end{itemize}

\begin{figure}[t]
  \centering
  \includegraphics[width=1.0\linewidth]{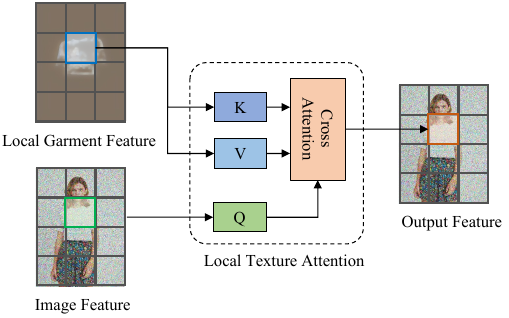}
  \caption{The Local Texture Attention mechanism. The local garment feature is extracted by masking the warped garment $G_w$ with the Auto Mask Module and is spatially aligned with the image feature. Those two features are then divided into non-overlap windows with the same partition method, and cross attention is applied only between the corresponding windows.}
  \label{fig:windowattention}
  
\end{figure}

\subsection{Pixel Aligned Feature Guidance}
Following the SD inpainting model, we concatenate the VAE encoded garment-agnostic image $I_a$ and the inpainting mask (resized according to the VAE spatial stride factor) to the noisy latent at each timestep. We also observed that adding two additional mask hints, the foreground mask and skin region mask, leads to more consistent visual quality. To prevent information leakage and deal with possible warping artifacts, we augment these mask regions by adding random dilation and erosion. 

A major difference between our method and previous methods~\cite{gou2023taming,morelli2023ladi} is that we do not directly introduce the warped garment $G_w$ by concatenating it to the noisy input of the diffusion UNet. Our observation is that concatenated features are typically assumed to be pixel-aligned with the diffusion target, which results in two primary issues: firstly, there is a clear mismatch between the warped garment $G_w$ and the garment from the ground-truth person image. When there is an error in predicting the deformation of the garment, this leads to degraded try-on results. 
Secondly, even when the predictions are accurate, the generated outcomes often retain all the information from the warped garment. The generator struggles to distinguish between wrinkles and textures. Consequently, when the clothing is too flat in the in-store image, it becomes challenging to produce realistic try-on result.

In this work, we instead propose a Auto Mask Module that takes the garment and disregards areas that are unrealistic or even erroneous.
Subsequently, we integrate the enhanced warped garment via a local cross attention mechanism (Sec.~\ref{sec:Garment_feature}), which allows us to address the challenge of incomplete alignment and unrealistic effects. 



\subsection{Garment Feature Guidance}\label{sec:Garment_feature}
\paragraph{Global Garment Feature}
To capture comprehensive information about the garment, such as garment style, color etc., our WarpDiffusion leverages a pretrained CLIP image encoder~\cite{radford2021learning} to extract features of in-shop garment $G$. The extracted feature vector is then compressed to the same dimensionality as the timestep embedding using an MLP and summed with it.


\paragraph{Local Texture Attention}\label{sec:attention}
The CLIP image embedding alone cannot capture all the fine details of the garment. To solve this, TryonDiffusion~\cite{zhu2023tryondiffusion} injects the un-warped garment in the cross attention layer.
However, since their cross attention layer is responsible for estimating the deformation field, it must calculate cross attention between the whole garment and the entire image, which greatly increases the computational complexity.
Instead, we use the warped garment as the input for the cross attention layer, enabling the model to focus solely on texture generation.
Since the warped garment is almost aligned in space, we do not need to calculate the cross attention between the warped garment and the whole image, and can instead follow the process outlined in Figure~\ref{fig:windowattention}. 
We first divide the original image features and warped garment features into windows through window partition. The attention is then only performed within each window region, reducing the overall computation considerably. After we perform the calculation window-by-window, we use the reverse operation to stack the windows back together and obtain the output. 
Note that when the window size is 1, this module is almost the same as concatenating, thus we use a relatively large window size (8 in our experiments, which is effectively similar to the prompt token length of 64).

\paragraph{Auto Mask Module}\label{sec:automask}
Even though we use the local texture attention to enhance the details of the try-on results, the model still lacks the ability to distinguish unrealistic or incorrect textures, resulting in unrealistic results. To address this, instead of retaining unrealistic and non-contributing elements produced by the warping module, our primary objective is to preserve only detailed and valuable textures. We therefore introduce the Auto Mask Module, which is designed to generate a mask over the informative regions. Specifically, we use a UNet to predict an informative mask $M_\text{info}$ from the VAE encoded warped garment $E(G_w)$ and the warped garment mask $M_w$.

We make several hypotheses about the generated mask. First, it should have a high masking ratio to retrain only the most informative feature, which is achieved by adding a $L_1$ regularization term, denoted as $L_{\text{min}}$. $L_{\text{min}}$ represents the mean value of all features of $M_{\text{info}}$. Secondly, we have observed that incorporating some prior knowledge into the mask contributes to the stability of training. For this purpose, we utilize a Laplacian filter to process the warped garment, obtaining a mask that serves as the ground truth, which is aimed at retraining regions with higher gradients (usually corresponding to textures such as logos).
Moreover, we calculate the Mean Squared Error (MSE) between the ground truth and the $M_\text{info}$, denoted as $L_{\text{preserve}}$. Lastly, the masked feature should be capable of recovering the original image. This is accomplished through the joint training of the auto mask UNet with the diffusion UNet.

\begin{figure*}[t]
  \centering
  \includegraphics[width=1.0\hsize]{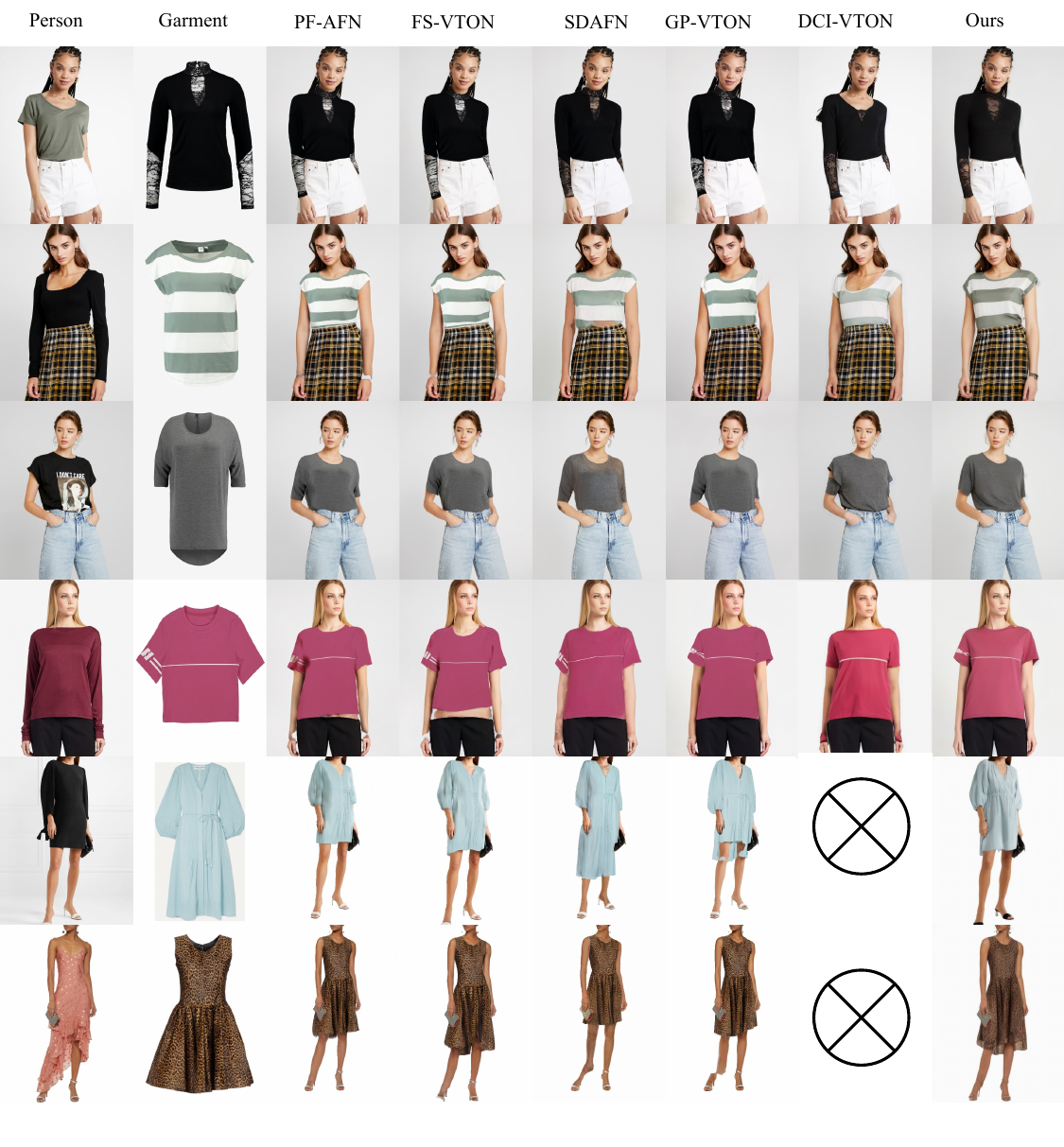}
    \vspace{-6mm}
  \caption{Qualitative comparisons on VITON-HD~\cite{choi2021viton}, DressCode~\cite{morelli2022dress}, and the in-the-wild test set. The first and second rows illustrate outcomes on VITON-HD~\cite{choi2021viton}, the third and fourth rows depict results on the in-the-wild scenario, and the last two rows showcase results on DressCode~\cite{morelli2022dress}. Our method consistently produces more realistic images. Note, the available DCI-VTON model does not support the generation of dresses.} 
  \vspace{-6mm}
\label{fig:Qualitative}
\end{figure*}




The final objective function for the Auto Mask Module is, 
\begin{equation}
    L_{\text{total}} = L_{\text{dm}}+ L_{\text{preserve}}+L_{\text{min}},
\end{equation}
where $L_{\text{dm}}$ represents the loss of the original diffusion model.
During training, we extract the ground truth garment from the person image and apply random elastic transformations and Gaussian blur to create the pseudo warped garment. The pixel-aligned features mentioned earlier are also derived from the ground truth target image by adding random noise. Additionally, to improve the model's recognition of real warp noise, we augment the training dataset with warped garments from baseline warping-based methods (e.g.,~\cite{xie2023gp}), integrating them in the final five epochs.

%% file: sec/4_results.tex
\section{Results}

\noindent \textbf{Datasets.}
We conduct extensive experiments on two public high-resolution VITON benchmarks, namely VITON-HD~\cite{choi2021viton} and DressCode~\cite{morelli2022dress}. For each dataset, we separately conduct experiments under the resolution of $512 \times 384$ and $1024 \times 768$.
Specifically, VITON-HD consists of 13,679 image pairs of front-view upper-body woman and upper-body in-store garment images, which are further split into 11,647/2,032 training/testing pairs. 
DressCode contains 48,392/5,400 training/testing pairs of front-view full-body person and in-store garment images, which are composed of three subsets with different category pairs (i.e., upper, lower, dresses). For in-the-wild testing, we additionally gathered images from the Internet that were not part of the above datasets. 

\noindent \textbf{Implementation Details.}
We employ SD-inpainting 2.0~\cite{SDInpainting} as our pretrained model. Our 512 resolution model is trained on 8 Tesla V100 GPUs, with a batch size of 6 for each GPU. The model is trained for 120 epochs with
learning rate 5e-5 over 3 days. For the 1024 resolution model, training is conducted on 8 NVIDIA A100 GPUs, configured with a batch size of 8 for each GPU. This model is trained for 100 epochs with a learning rate of 1e-5 over 1 day.

\begin{figure}[t]
  \centering
  \includegraphics[width=1.0\hsize]{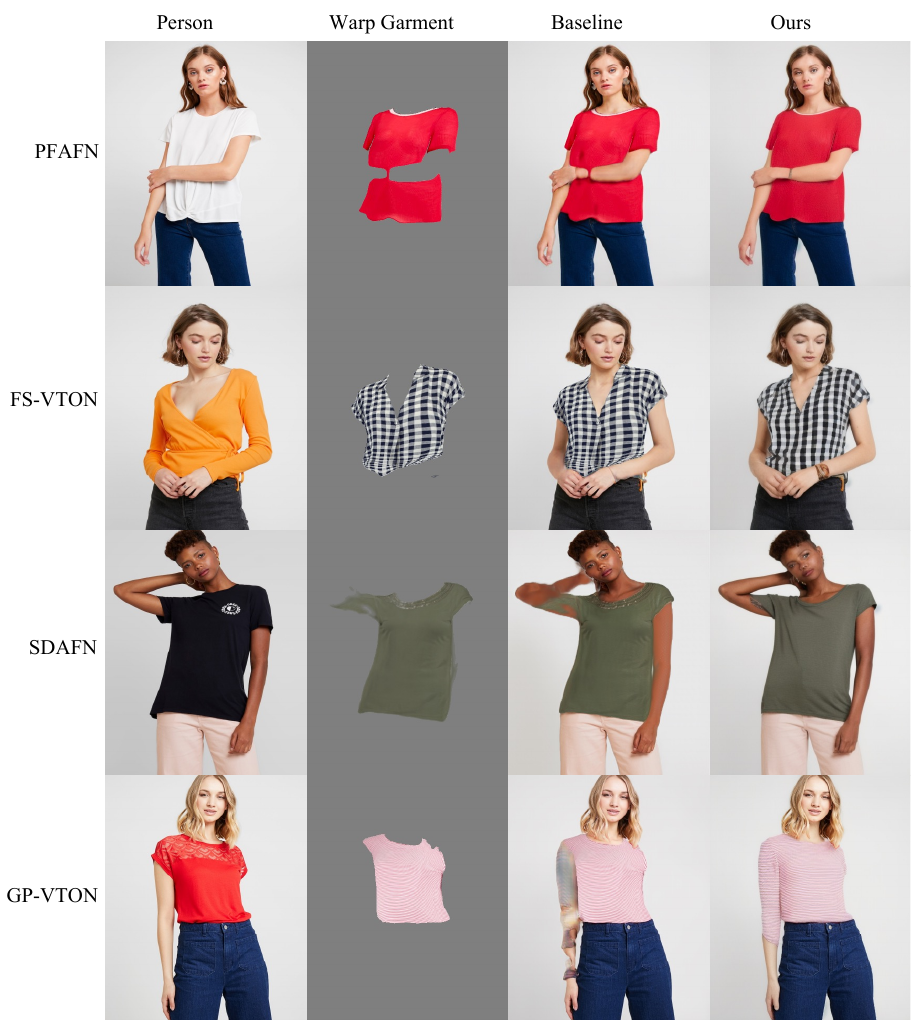}
  \caption{Qualitative comparisons to warping-based approaches when our model uses their warped garment in a plug-and-play manner. WarpDiffusion consistently improves the baseline results.
  }
\label{fig:different_warp}
\vspace{-6mm}
\end{figure}

\begin{table*}[t]

\def\arraystretch{1.2}
\small
\tabcolsep 8pt

\centering
 \vspace{-2mm}

 \vspace{-5mm}

\begin{tabular}{c c c c c c c c c c c c c c c}
  \toprule

   \multicolumn{2}{c}{Resolution} & & &\multicolumn{4}{c}{$512 \times 384$} & & &\multicolumn{4}{c}{$1024 \times 768$}  \\

    \cmidrule{1-2} \cmidrule{5-8} \cmidrule{11-14} 
  
  \multicolumn{2}{c}{Method}                              
  & & &SSIM $\uparrow$ & FID $\downarrow$ & LPIPS $\downarrow$ & HE $\uparrow$ & & &SSIM $\uparrow$ & FID $\downarrow$ & LPIPS $\downarrow$ & HE $\uparrow$  \\
    \cmidrule{1-2} \cmidrule{5-8} \cmidrule{11-14} 
  \multicolumn{2}{c}{PF-AFN~\cite{ge2021parser} }
  & & &0.885 & 9.616 & 0.087 & 0.06 &  & &0.898 & 9.807 & 0.096& 0.03 \\
 \multicolumn{2}{c}{FS-VTON~\cite{he2022style}} 
  & & &0.881 & 9.735 & 0.091 &0.09 &  & &0.896 & 9.667 & 0.097&0.05 \\
  \multicolumn{2}{c}{SDAFN~\cite{bai2022single}} 
  & & &0.881 & 9.497 & 0.092 &0.04 &  & &0.892 & 9.783 & 0.108 &0.12 \\
  \multicolumn{2}{c}{GP-VTON~\cite{xie2023gp}} 
  & & &0.893 & 9.405 & 0.079 &0.11 &  & &0.898 & 9.238 & 0.091&0.14  \\
    \cmidrule{1-2} \cmidrule{5-8} \cmidrule{11-14} 
 \multicolumn{2}{c}{DCI-VTON~\cite{gou2023taming}} 
  & & &0.868 & 9.166 & 0.096 &0.12 &  & &- & - & - & - \\
    \cmidrule{1-2} \cmidrule{5-8} \cmidrule{11-14} 
  \multicolumn{2}{c}{\textbf{WarpDiffusion}} 
  & & &\textbf{0.896} & \textbf{8.902} & \textbf{0.078} & \textbf{0.58} & & &\textbf{0.901} & \textbf{9.187} & \textbf{0.089}& \textbf{0.66} \\  
  \bottomrule

\end{tabular}
\vspace{-2mm}
\caption{Quantitative comparisons at {$512 \times 384$} and {$1024 \times 768$} resolution on the VITON-HD dataset~\cite{choi2021viton}.}
\vspace{-4mm}
\label{tab:vitonhd_results}
\end{table*}

\noindent\textbf{Baselines and Evaluation Metrics.}
We perform quantitative and qualitative comparisons with four advanced GAN-based methods, namely PF-AFN~\cite{ge2021parser}, FS-VTON~\cite{he2022style}, SDAFN~\cite{bai2022single} and GP-VTON~\cite{xie2023gp}, and the latest Diffusion-based method, DCI-VTON~\cite{gou2023taming}. The GAN-based methods are trained from scratch on VITON-HD~\cite{choi2021viton} and DressCode~\cite{morelli2022dress} while for the diffusion-based model~\cite{gou2023taming}, we directly use their released pretrained model. We are not able to conduct extensive comparison with TryonDiffusion~\cite{zhu2023tryondiffusion} as it is not publicly available now. For all baselines, we strictly follow the official instructions to run the training and testing scripts.
Following previous work, we employ three widely used metrics (i.e., Structural SIMilarity index (SSIM)~\cite{wang2004image}, Perceptual distance (LPIPS)~\cite{zhang2018unreasonable}, and Fr$\mathbf{\acute{e}}$chet Inception Distance (FID)~\cite{parmar2022aliased}) to evaluate the similarity between the synthesized and real images. Furthermore, we conduct a Human Evaluation (HE) study to evaluate the synthesis quality of different methods\footnote{More details on the Human Evaluation are included in the supplementary material.}.
It should be noted that, unless otherwise stated, we leverage GP-VTON's warping module in WarpDiffusion. However, to illustrate WarpDiffusion's plug-and-play nature, we do provide results with alternative warping strategies in Fig.~\ref{fig:different_warp}.

\subsection{Comparison with the State-of-the-Art Methods}

\begin{table}[t]

\def\arraystretch{1.2}
\small
\tabcolsep 6pt

\centering

\begin{tabular}{c c c c c c c }
  \toprule

   \multicolumn{2}{c}{Resolution} & & \multicolumn{3}{c}{$512 \times 384$}  \\

    \cmidrule{1-2} \cmidrule{4-7} 
  
  \multicolumn{2}{c}{Method}                              
  & & SSIM $\uparrow$ & FID $\downarrow$ & LPIPS $\downarrow$ & HE $\uparrow$ \\
  \cmidrule{1-2} \cmidrule{4-7}  
  \multicolumn{2}{c}{PF-AFN~\cite{ge2021parser}} 
  & & 0.884 & 10.19 & 0.107& 0.07  \\
  \multicolumn{2}{c}{FS-VTON~\cite{he2022style}} 
  & & 0.886 & 9.20 & 0.099  & 0.08\\
  \multicolumn{2}{c}{SDAFN~\cite{bai2022single}} 
  & &0.892 & 8.69 & 0.089  & 0.11\\
  \multicolumn{2}{c}{GP-VTON~\cite{xie2023gp}} 
  & & 0.894 & 8.71 & 0.091 & 0.15 \\
  \cmidrule{1-2} \cmidrule{4-7}
  \multicolumn{2}{c}{\textbf{WarpDiffusion}} 
  & & \textbf{0.895} & \textbf{8.61} & \textbf{0.088} & \textbf{0.59} \\  
  \bottomrule

\end{tabular}
\caption{Quantitative comparisons at {$512 \times 384$} resolution on the DressCode dataset~\cite{morelli2022dress}.}
\vspace{-4mm}
\label{tab:dresscode}
\end{table}

\noindent\textbf{Quantitative Results} As reported in Tab.~\ref{tab:vitonhd_results} and Tab.~\ref{tab:dresscode} our WarpDiffusion consistently surpasses the baselines on all metrics for the VITON-HD~\cite{choi2021viton} and DressCode datasets~\cite{morelli2022dress}, demonstrating that WarpDiffusion can obtain more precise warped garments and generate try-on results with better visual quality. In addition, we designed a human evaluation to compare the
baselines and our generation results. A higher human evaluation
score indicates that a larger fraction of participants preferred the
results of a given method. As reported in Tab.~\ref{tab:vitonhd_results} and Tab.~\ref{tab:dresscode}, our proposed WarpDiffusion
model outperforms the baselines in most cases by a large margin. This means that our method can generate better and more realistic results also when judged by humans.

\noindent\textbf{Qualitative Results.}
Fig.~\ref{fig:Qualitative} provides the qualitative comparison of WarpDiffusion with the state-of-the-art baselines on the VITON-HD~\cite{choi2021viton} and DressCode datasets~\cite{morelli2022dress} as well as for the in-the-wild test set. The results demonstrate the superiority of our WarpDiffusion over the baselines.\footnote{More results are included in the supplementary material}
First of all, all of baselines fail to generate realistic wrinkles. In addition, compared with the diffusion base model, our method uses richer features which makes our texture consistency better than DCI-VTON~\cite{gou2023taming}. Finally, Fig.~\ref{fig:different_warp} 
shows the comparison between the results of our method and the original method using the same warping schemes. It shows that our method introduces robustness to excessive deformations and noisy warping results to a certain extent and generates more realistic results. This demonstrates the superiority of our approach and the benefit of its plug-and-play nature.

\begin{figure}
  \centering
  \includegraphics[width=1.0\hsize]{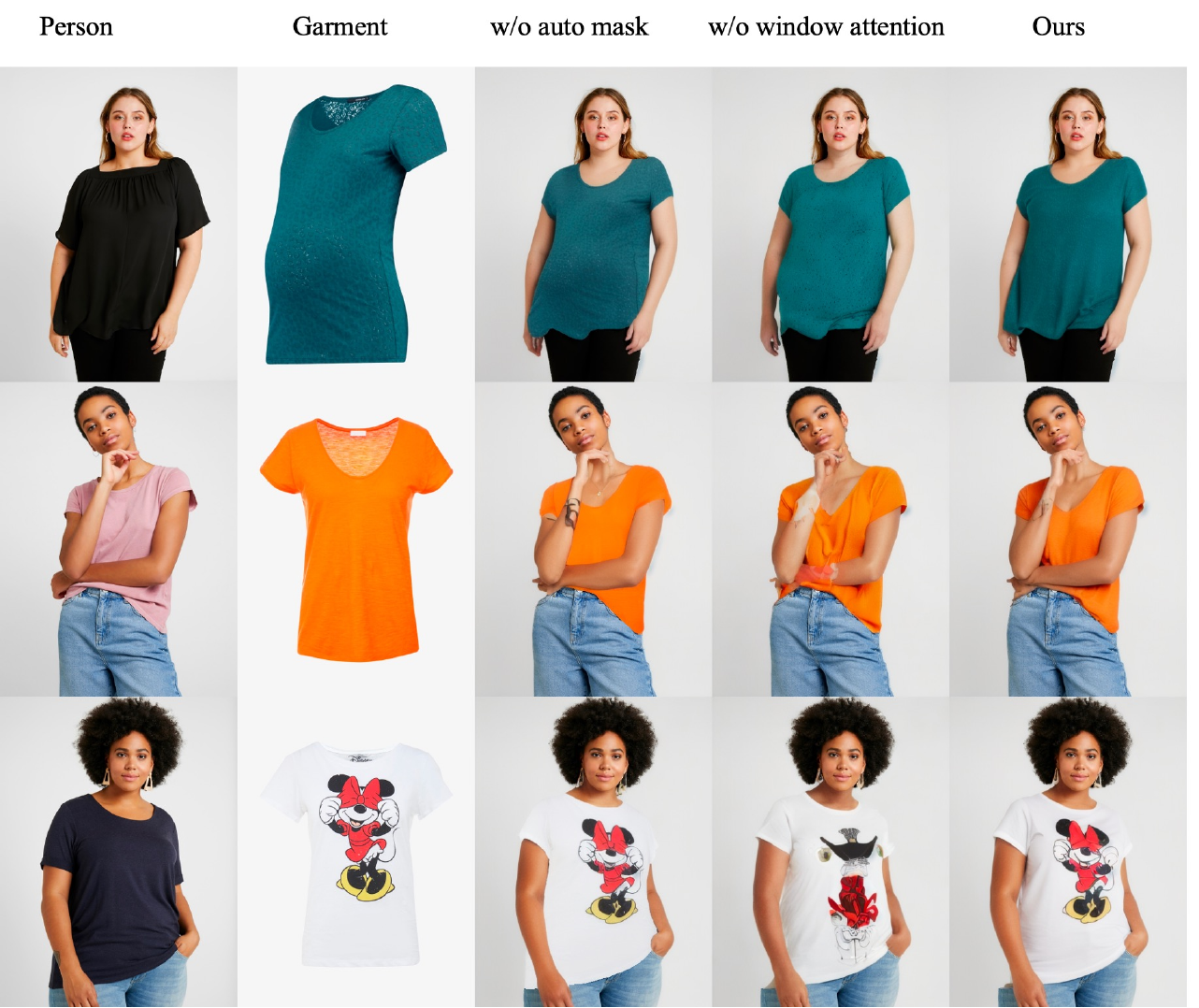}
  \vspace{-4mm}
  \caption{Qualitative ablation study highlighting the value of our Auto Mask Module and Local Texture Attention. When Auto Mask is not applied, the try-on results will retain unrealistic wrinkles from the original garment while the full model fits the garment realistically to the person.} 
  \label{fig:ablation}
  \vspace{-4mm}
\end{figure}

\subsection{Ablation Study}
To validate the effectiveness of the proposed Auto Mask Module, we conducted an ablation study by comparing WarpDiffusion with a version that does not include the auto mask module. 
The comparison was performed on the VITON-HD dataset~\cite{choi2021viton}, and we observed that the full WarpDiffusion model achieved improved metric scores. This indicates that the generated results are more aligned with the real distribution of the dataset, thus validating the impact of auto mask in preserving significant information while improving the ability to generate other areas.

To further validate the effectiveness of the local texture attention, we conducted another ablation study. In this ablation study, we compared WarpDiffusion with versions that only utilized the CLIP feature of the unwarped garment and removed the local texture attention. Once again, we compared the main metric scores and observed that the scores worsened when the local texture attention was removed. This indicates that the rich garment features play a significant role in the generation process.

We also provide qualitative results in Fig~\ref{fig:ablation} of these ablation studies, which further demonstrate the benefits of our proposed method. Specifically, we observed that removing the auto mask module weakened the ability to generate new wrinkles and address excessive distortion. The model primarily preserved the existing wrinkles from the warped garment, resulting in less realistic results. Furthermore, when we eliminated the local texture attention, we notice a significant reduction in texture consistency.

\subsection{Visualization of Local Garment Feature}
\begin{figure}[t]
  \centering
  \includegraphics[width=1.0\hsize]{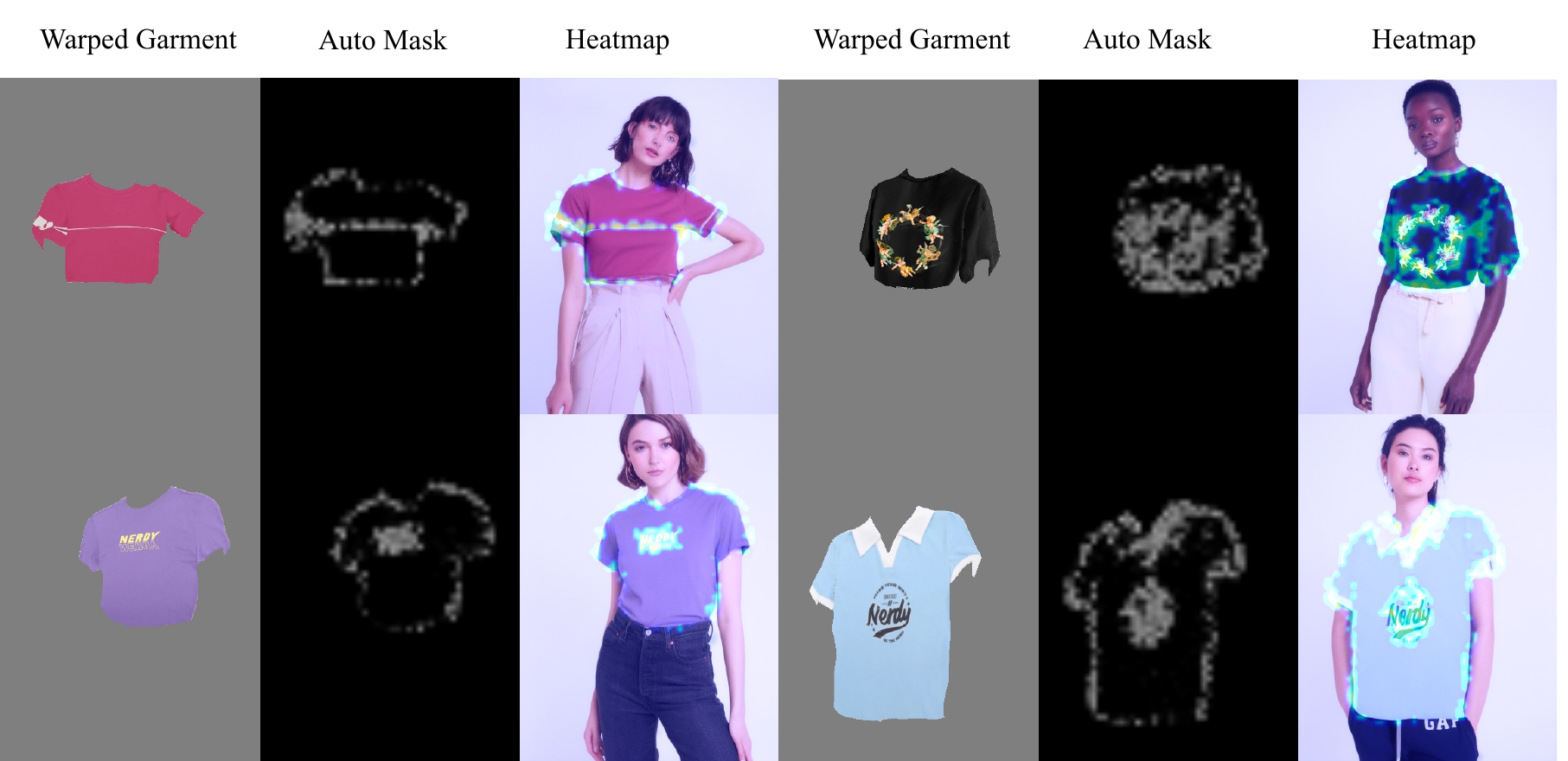}
  \caption{Visualization of the auto mask for warped garments.}
\label{fig:automask}
\vspace{-3mm}
\end{figure}

In this section, we further demonstrate the remarkable ability of our proposed Auto Mask Module to capture fine-grained texture. To provide a clear illustration, we employ heatmaps to highlight the specific areas that the Auto Mask Module focuses on. As depicted in Figure~\ref{fig:automask}, the Auto Mask Module effectively extracts the required information while intelligently disregarding wrinkles and other insignificant features. This ensures that regions without significant textures will be enhanced by a more powerful generator, leading to more realistic results, while at the same time preserving the overall texture consistency.

\begin{table}[t]
\def\arraystretch{1.2}
\small
\tabcolsep 6pt
\centering

\begin{tabular}{c c c c c c }
  \toprule
    \multicolumn{2}{c}{Method}                              
  & & SSIM $\uparrow$ & FID $\downarrow$ & LPIPS $\downarrow$ \\
  \cmidrule{1-2} \cmidrule{4-6}  
  \multicolumn{2}{c}{w/o Auto Mask Module} 
  & & 0.894 & 9.10 & 0.079   \\
  \multicolumn{2}{c}{w/o Local Texture Attention } 
  & & 0.831 & 9.23 &  0.102 \\
  \cmidrule{1-2} \cmidrule{4-6} 
  \multicolumn{2}{c}{\textbf{WarpDiffusion}} 
  & & \textbf{0.896} & \textbf{8.90} & \textbf{0.078} \\  
  \bottomrule
\end{tabular}
\caption{Ablation results on the VITON-HD dataset~\cite{choi2021viton} demonstrating the benefit of the proposed Auto Mask Module and Local Texture Attention.}
\vspace{-2mm}
\label{tab:ablation}
\end{table}

\section{Conclusion}


In this work, we introduce WarpDiffusion, a plug-and-play model to generate high-fidelity VITON results with authentic realistic try-on effects. To optimize resource utilization and enhance synthesis quality, WarpDiffusion integrates a pre-trained StableDiffusion model along with a novel local texture attention mechanism. In order to attain more realistic try-on effects, WarpDiffusion introduces the Auto Mask Module to refine the warped garment. Experiments conducted on high-resolution VITON benchmarks and a diverse in-the-wild test set affirm WarpDiffusion's superior efficacy compared to existing methods. 

\noindent\textbf{Limitation and future work.} 



A limitation of our approach is that when faced with very complex poses and more sophisticated try-on styles (such as layering, inner wear), the quality of the garment warping is not sufficient, leading to sub-optimal results. Besides, we use the stable diffusion backbone with lossy compression VAE, thus can not fully preserve all high-frequency details. To address these limitations, we aim t explore how to control the try-on process in a more fine-grained manner.

%% file: sec/supp_arxiv.tex
\clearpage
\setcounter{page}{1}
\maketitlesupplementary

\section{Introduction}

In this supplementary material, we provide additional details of the architecture (Sec.~\ref{sec:AD}) and the human evaluation (Sec.~\ref{sec:HE}). Further, we include additional qualitative results of our method (Sec.~\ref{sec:AR}) and ablated results (Sec.~\ref{sec:AS}). Finally, we provide an analysis of the potential social impact and an analysis of failure cases (Sec.~\ref{sec:fcase}).

\section{Architecture Details}\label{sec:AD}

The structure of WarpDiffusion resembles Stable Diffusion~\cite{SDInpainting}, with the main differences lying in its acceptance of an 11-channel input. This input includes 4 channels of noise, 4 channels of garment-agnostic representations, 1 channel corresponding to the mask, and one channel each for the foreground mask and skin mask. Additionally, in the cross-attention layer, it takes as input the 4 channels of the warped garment enhanced by the auto-mask module and its corresponding mask.

As for the auto-mask module, it leverages a three-layer Feature Pyramid Network (FPN). Each layer consists of a convolution with a stride of 2, succeeded by two residual blocks.

\section{Human Evaluation Details}\label{sec:HE}

\begin{figure*}
  \centering
  \includegraphics[width=1.0\hsize]{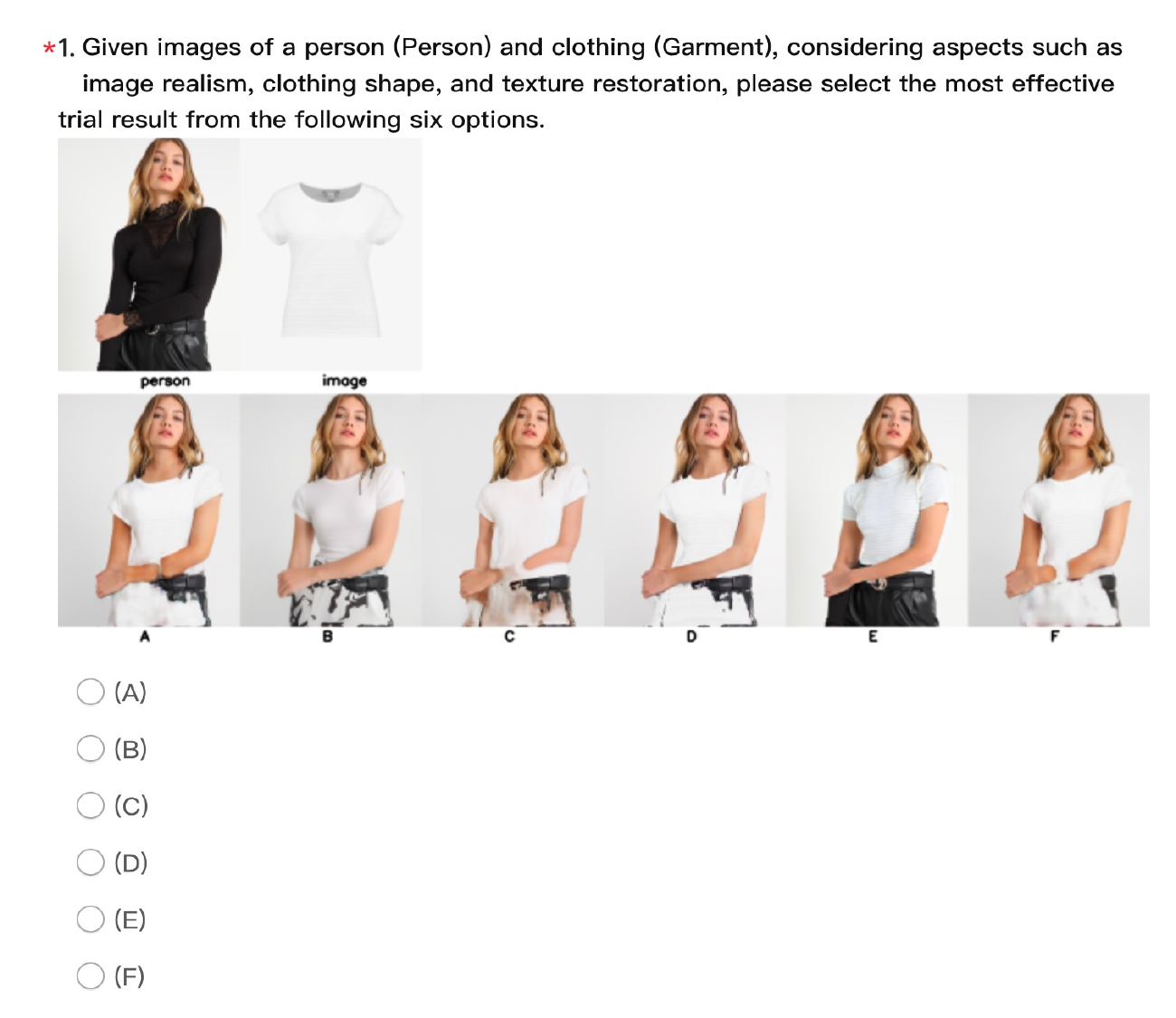}
  \caption{Interface of the questionnaire used to evaluate the final try-on results on the VITON-HD dataset~\cite{choi2021viton}.} 
  \label{fig:human}

\end{figure*}

For the human evaluation, we separately designed two questionnaires for the VITON-HD dataset~\cite{choi2021viton} and the DressCode dataset~\cite{morelli2022dress}. Specifically, for the VITON-HD dataset, 100 volunteers are invited to complete the questionnaire which is composed of 25 assignments. For the DressCode dataset, 100 volunteers are invited to complete the questionnaire which contains 15 assignments for each garment category (i.e., upper-, lower-garment, dresses), namely, 45 assignments in total. For each assignment in the questionnaire, given a person image and a garment image, the volunteers are asked to select the most realistic and accurate try-on result out of six options, which are generated by our WarpDiffusion and the baseline methods (i.e., PF-AFN~\cite{ge2021parser}, FS-VTON~\cite{he2022style}, SDAFN~\cite{bai2022single}, GP-VTON~\cite{xie2023gp}, DCI-VTON~\cite{gou2023taming}). \
The order of the generated results in each assignment is randomly shuffled. Fig.~\ref{fig:human} shows the interface of the questionnaire for the VITON-HD dataset. The interface for the DressCode dataset is identical. 

\section{Additional Results}\label{sec:AR}

\begin{figure*}[t]
  \centering
  \includegraphics[width=1.0\hsize]{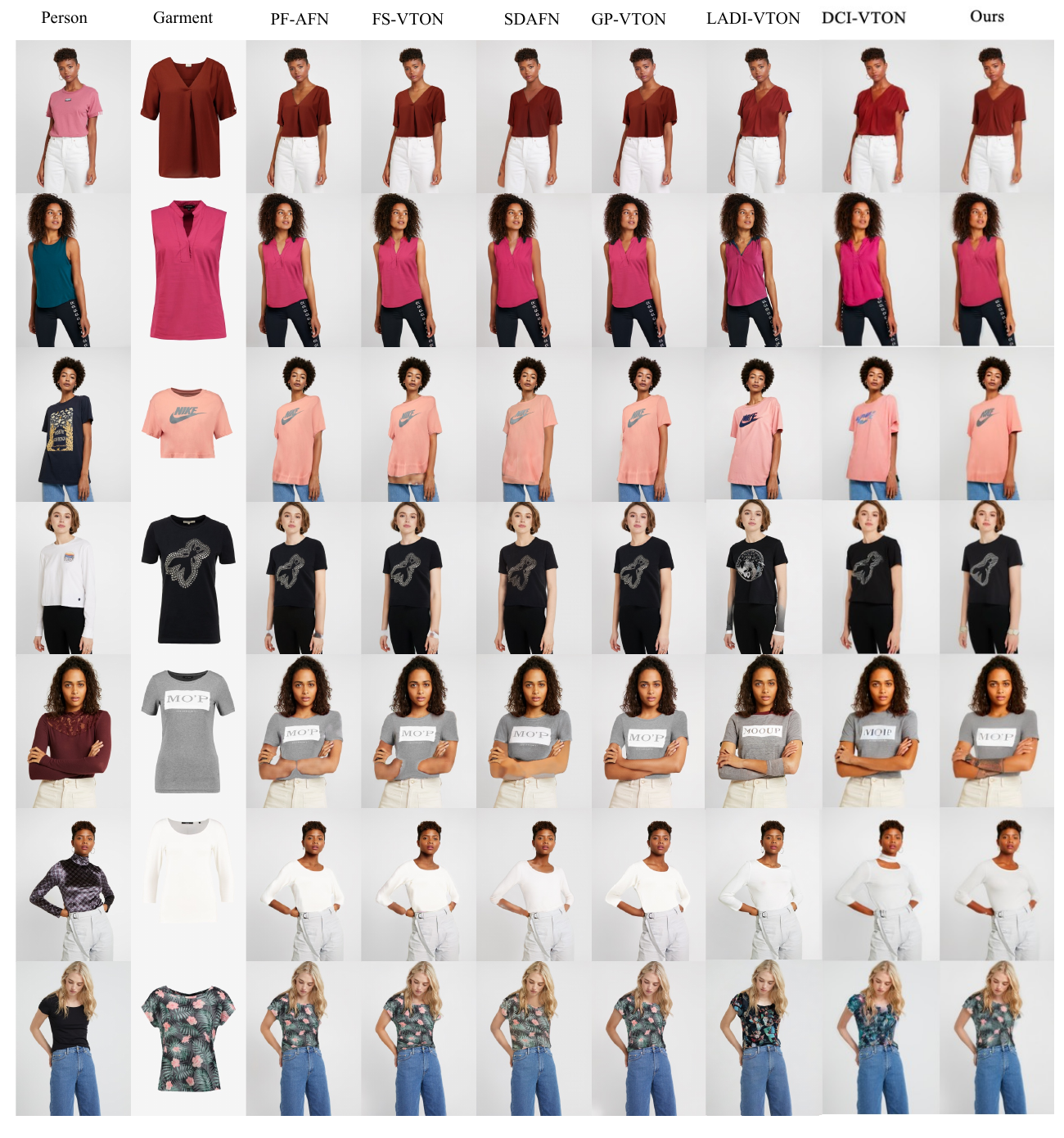}
  \caption{Qualitative comparisons on VITON-HD~\cite{choi2021viton}. Please zoom in for more details.} 
\label{fig:suppviton}
\end{figure*}

\noindent\textbf{Visual Comparisons with SOTAs on the VITON-HD dataset~\cite{choi2021viton}}.
Fig.~\ref{fig:suppviton} displays additional visual comparisons of WarpDiffusion with the baseline methods on the VITON-HD dataset. 

\begin{figure*}[t]
  \centering
  \includegraphics[width=1.0\hsize]{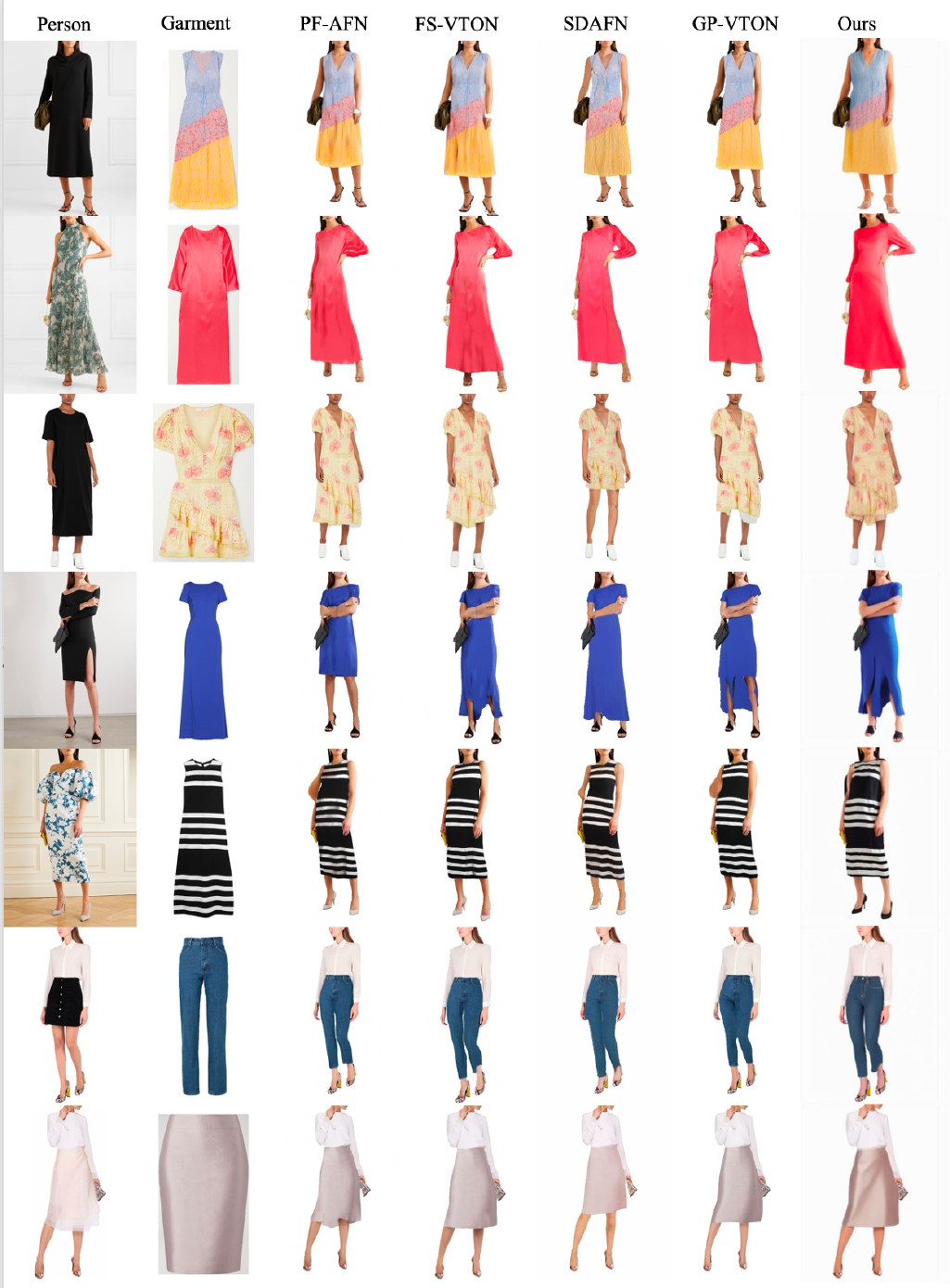}
  \caption{Qualitative comparisons on DressCode~\cite{morelli2022dress}. Please zoom in for more details.} 
\label{fig:suppdress}
\end{figure*}

\noindent\textbf{Visual Comparisons with SOTAs on the DressCode dataset~\cite{morelli2022dress}}.
Fig.~\ref{fig:suppdress} displays additional visual comparisons of WarpDiffusion 
with the baseline methods on the DressCode dataset.

\begin{figure*}[t]
  \centering
  \includegraphics[width=1.0\hsize]{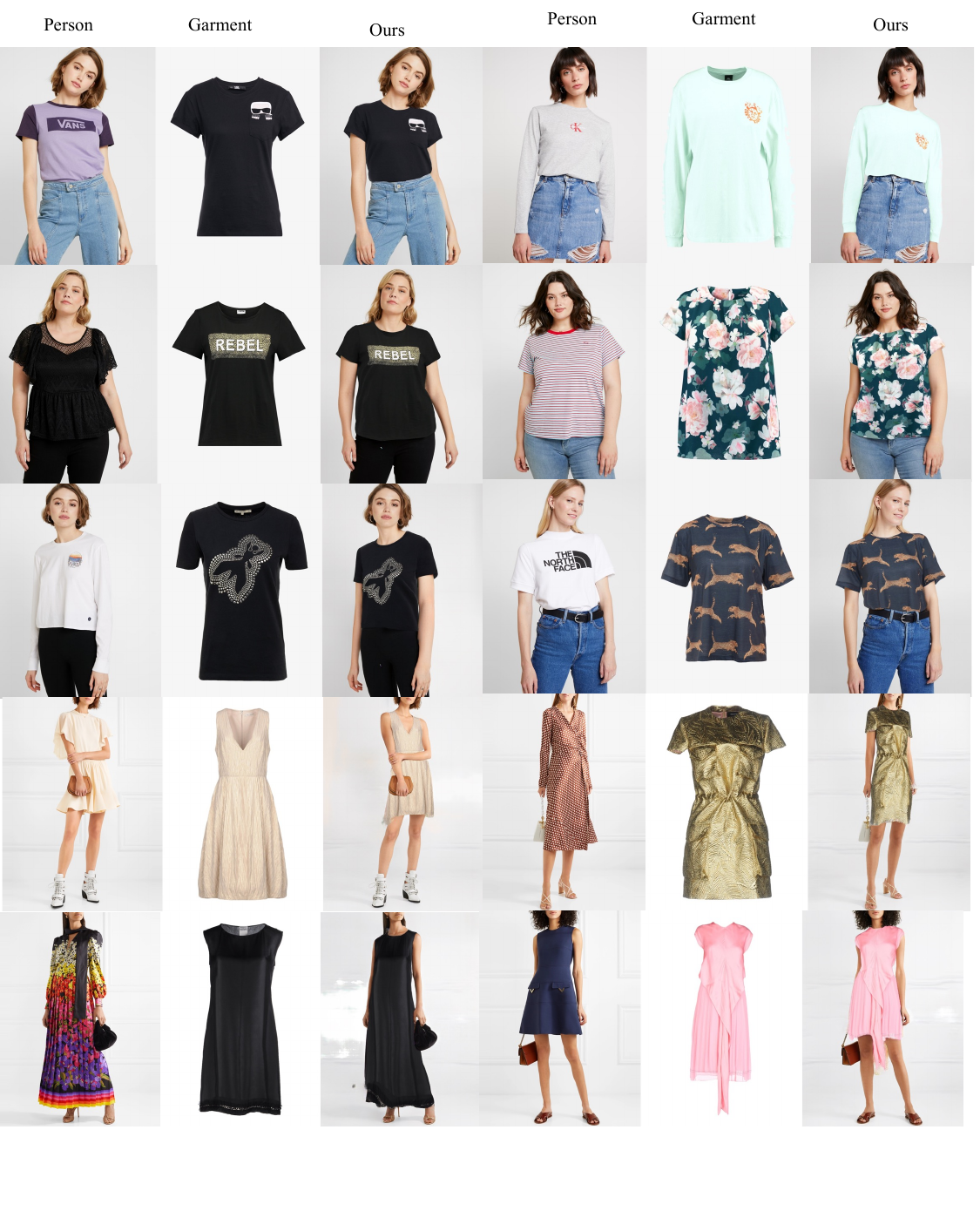}
  \caption{Qualitative results at 1024 resolution. Please zoom in for more details.} 
\label{fig:supp1024}
\end{figure*}

\noindent\textbf{Virtual Try-on for High Resolution}. 
Fig.~\ref{fig:supp1024} displays additional results of WarpDiffusion at the resolution of 1024.

\section{Additional Ablation Study}\label{sec:AS}
\begin{figure}
  \centering
  \includegraphics[width=1.0\hsize]{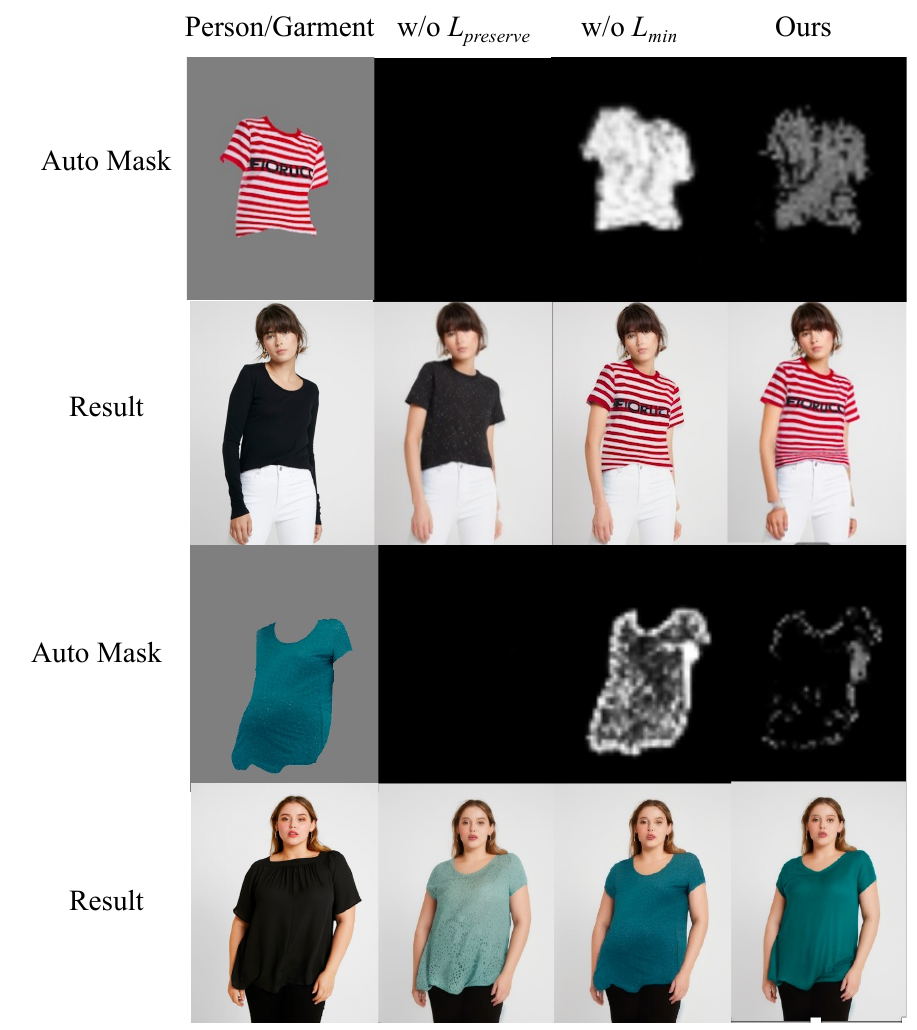}
  \caption{
  Results for the Auto Mask Module ablation study.
  } 
  \label{fig:a_loss}

\end{figure}

To validate the effectiveness of the loss function in the Auto Mask Module, we conducted an ablation experiment specifically focusing on this module. We generated two versions of WarpDiffusion, where the first version does not include $L_{\text{min}}$, while the second version does not include $L_{\text{preserve}}$. Qualitative results are provided in Fig~\ref{fig:a_loss}. In the absence of  $L_{\text{min}}$, the generated mask often retains a significant amount of information, leading to the preservation of undesired false wrinkles. On the other hand, without $L_{\text{preserve}}$, the generated mask retained very little information, resulting in the inability to generate the correct try-on results.

\begin{figure}
  \centering
  \includegraphics[width=1.0\hsize]{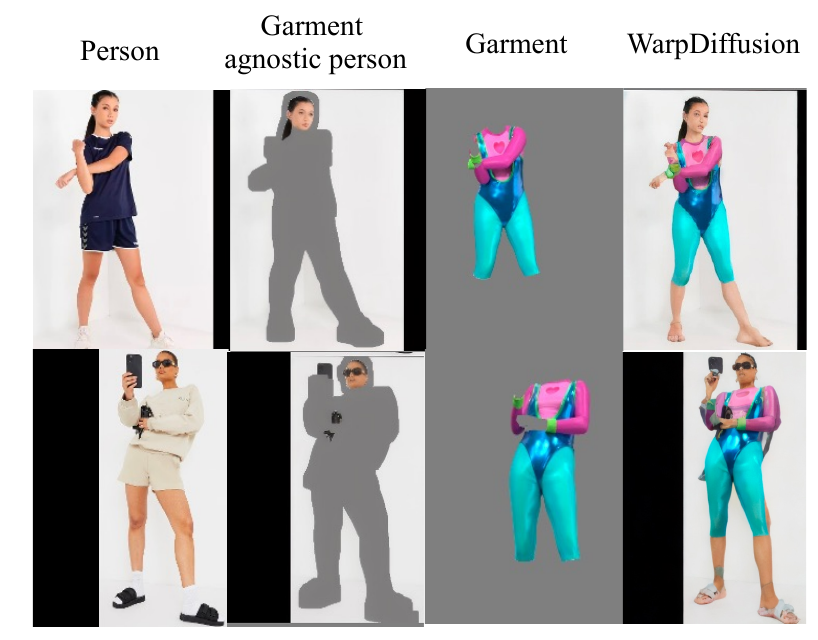}
  \caption{Failure cases of our WarpDiffusion.}
    \label{fig:failure_case}
\end{figure}

\section{Potential Social Impacts and Limitations}\label{sec:fcase}

\noindent\textbf{Potential social impacts.}
As with most generative approaches, WarpDiffusion can be used for malicious purposes by generating images that infringe upon copyrights and/or privacy. Given these considerations, responsible use of the model is advocated.

\noindent\textbf{Limitations.}
 As shown in Fig.~\ref{fig:failure_case}, our WarpDiffusion encounters challenges when faced with difficult poses and unfamiliar garments, making it difficult to generate satisfactory try-on results. In order to address this issue, we are exploring ways to enhance the model's zero-shot generation capability and achieve finer control.

